\documentclass{article}

\usepackage[preprint, nonatbib]{neurips_2020}

\usepackage[utf8]{inputenc} 
\usepackage{hyperref}       
\usepackage{times}
\usepackage{soul}
\usepackage{url}
\usepackage{graphicx}
\usepackage{amsmath}
\usepackage{amsthm}
\usepackage{booktabs}
\usepackage{algorithm}
\usepackage{enumitem}

\usepackage[T1]{fontenc}    
\usepackage{amsfonts}       
\usepackage{nicefrac}       
\usepackage{microtype}      
\usepackage{authblk}

\usepackage{multirow}
\usepackage{float}
\usepackage{subfig}

\DeclareMathOperator*{\argmin}{arg\,min}

\usepackage{color}
\usepackage{wrapfig}
\usepackage{caption}

\usepackage{algpseudocode}
\usepackage{algorithmicx}

\usepackage{tikz}
\usetikzlibrary{shapes,arrows}
\usetikzlibrary{external}
\usetikzlibrary{positioning,calc}
\tikzset{%
  block/.style    = {draw, thick, rectangle, minimum height = 3em,
    minimum width = 3em},
  sum/.style      = {draw, circle, node distance = 2cm}, 
  input/.style    = {coordinate}, 
  output/.style   = {coordinate} 
}

\usepackage{microtype}      
\usepackage{doi}

\title{Cross-domain Cross-architecture Black-box Attacks on Fine-tuned Models with Transferred Evolutionary Strategies}

\author[1]{Yinghua Zhang
}
\author[1]{Yangqiu Song
}
\author[2]{Kun Bai
}
\author[1,3]{Qiang Yang
}
\affil[1]{Hong Kong University of Science and Technology \authorcr
  \{\tt yzhangdx, yqsong, qyang\}@cse.ust.hk}
\affil[2]{IEEE Senior Member \authorcr
  \tt kunbai@ieee.org}
\affil[3]{WeBank}

\date{}

\begin{document}
\maketitle

\begin{abstract}
Fine-tuning can be vulnerable to adversarial attacks.
Existing works about black-box attacks on fine-tuned models (BAFT) are limited by strong assumptions. To fill the gap, we propose two novel BAFT settings, cross-domain and cross-domain cross-architecture BAFT, which only assume that (1) the target model for attacking is a fine-tuned model, and (2) the source domain data is known and accessible. 
To successfully attack fine-tuned models under both settings, we propose to first train an adversarial generator against the source model, which adopts an encoder-decoder architecture and maps a clean input to an adversarial example. Then we search in the low-dimensional latent space produced by the encoder of the adversarial generator. The search is conducted under the guidance of the surrogate gradient obtained from the source model. Experimental results on different domains and different network architectures demonstrate that the proposed attack method can effectively and efficiently attack the fine-tuned models. 
\end{abstract}


\section{Introduction}
Fine-tuning is one of the most popular transfer learning methods for training deep learning models with limited labeled data. 
Many deep learning frameworks, including TensorFlow, PyTorch, MxNet, release pre-trained models to the public, and users can reduce the data labeling burden and save computational costs by fine-tuning from the pre-trained models. 
However, fine-tuned models are vulnerable to adversarial attacks~\cite{wang_attackTL_usenix_2018,zhang_attackTL_kdd_2020}. An adversarial example is an imperceptible perturbation to the original input so that a target model will make an erroneous prediction \cite{szegedy_iclr_2013}. Based on the availability of the knowledge about the target model, adversarial attacks can be categorized into two types, namely \emph{white-box} and \emph{black-box} attacks. Black-box attacks do not require the knowledge about the training data and model parameters, and they can only make a few or no queries to the target model, which is a more practical setting. Pioneering works \cite{wang_attackTL_usenix_2018,zhang_attackTL_kdd_2020} demonstrate that fine-tuned models can be attacked in a black-box manner. 

\begin{figure*} 
\centering
    \begin{minipage}{0.48\linewidth}
    \centering
    \subfloat[Vanilla black-box attacks] {
    \centering
    \includegraphics[scale=0.27]{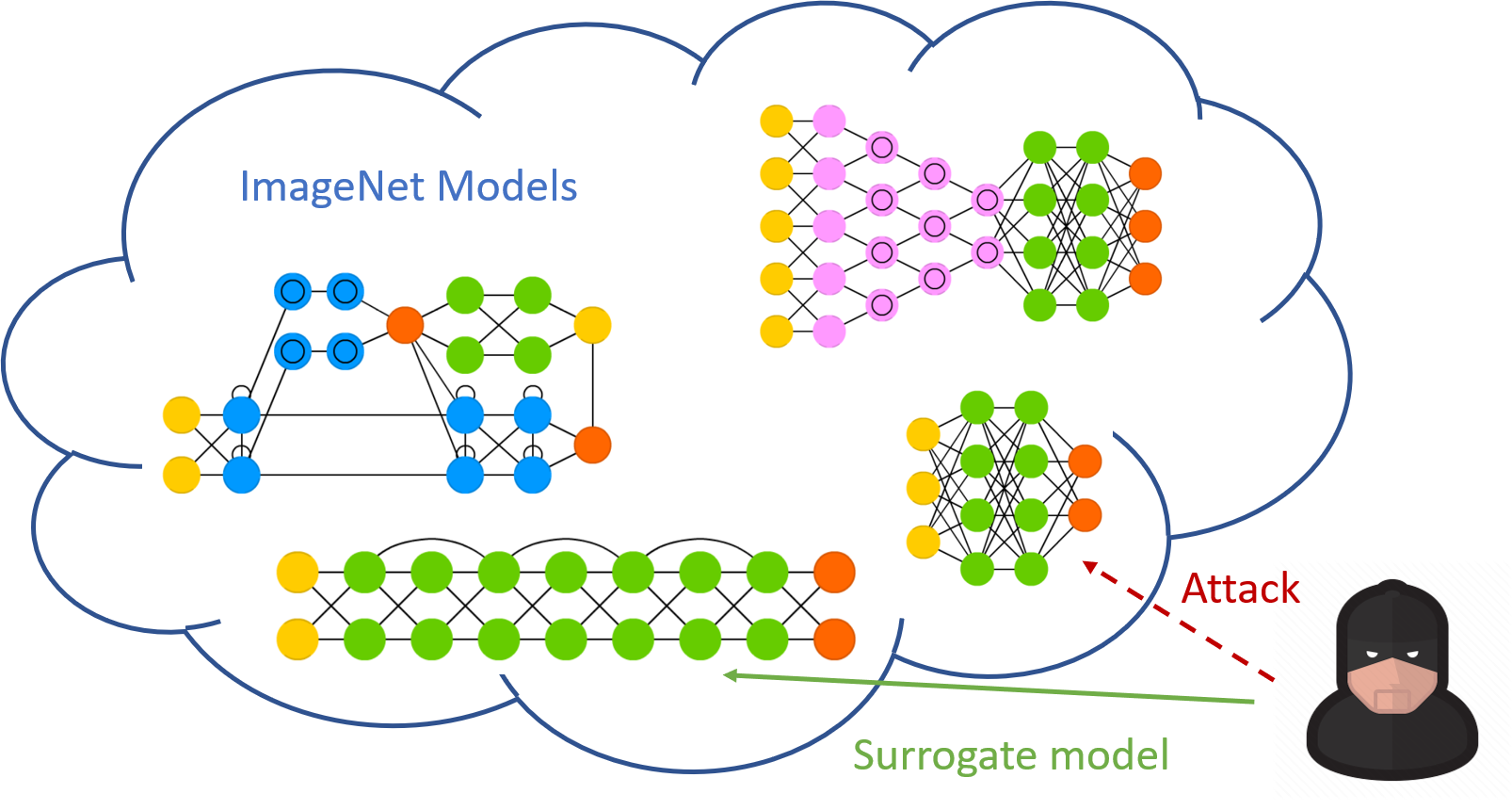}
    }
    \end{minipage}
    \hfill
    \begin{minipage}{0.48\linewidth}
    \centering
    \subfloat[Black-box attacks on fine-tuned models (BAFT)] {
    \centering
    \includegraphics[scale=0.25]{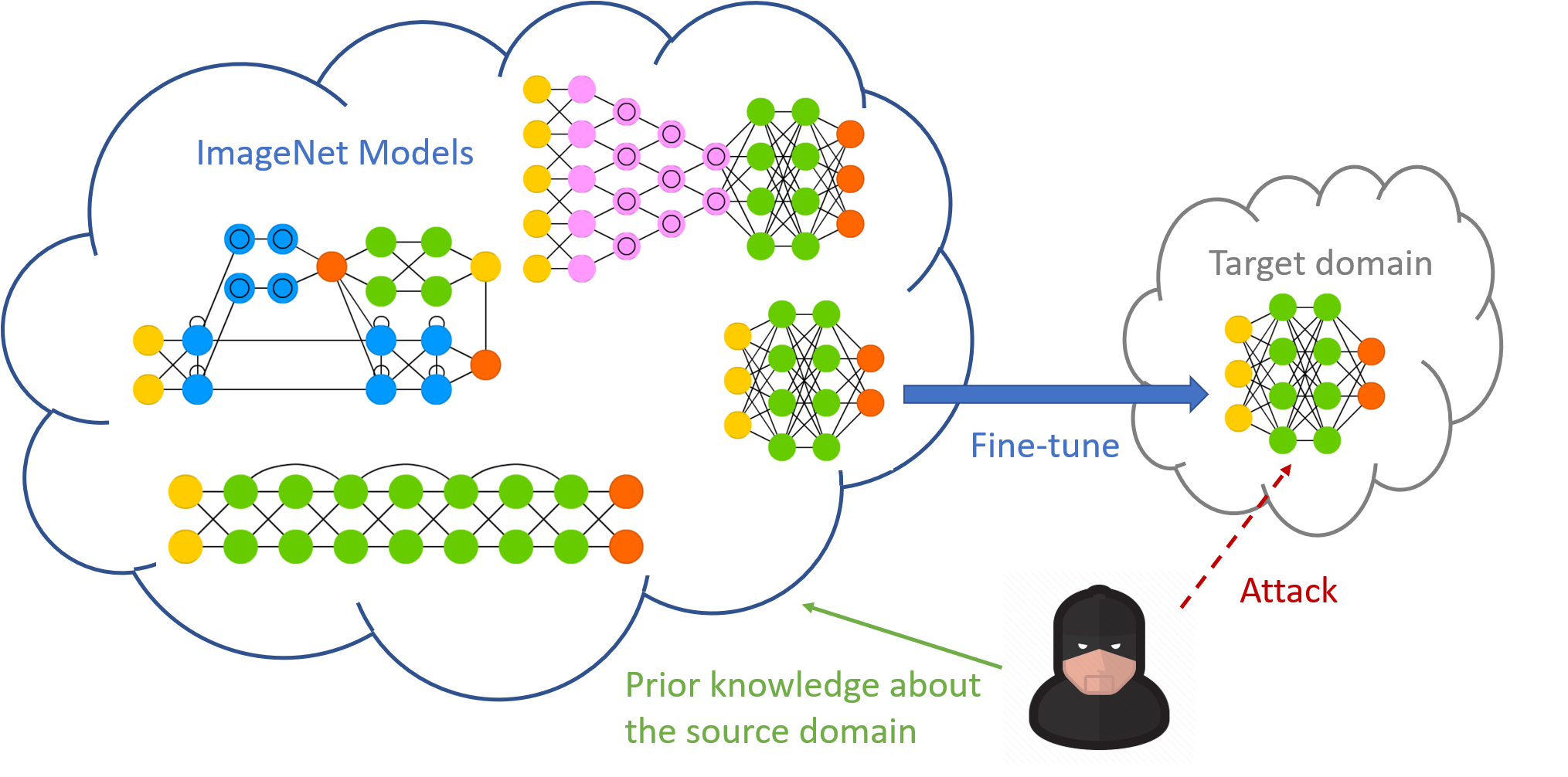}
    }
    \end{minipage}
    \caption{
    The difference between the vanilla black-box attack setting and the proposed BAFT settings. For vanilla black-box attacks, we may use a substitute model (\texttt{Model} \texttt{A}) to attack a target model (\texttt{Model} \texttt{B}), but both models are trained within the same domain. In the BAFT settings, the target model is fine-tuned from a source model that is pre-trained on a giant domain, which has become a de-facto paradigm for deep learning with limited data. It is expected to craft effective and efficient adversarial attacks by leveraging the prior knowledge about the source domain. Yet it remains challenging due to non-trivial discrepancies between the two domains. 
    }
    \label{fig:schema-baft}
\end{figure*}

Existing works about \underline{B}lack-box \underline{A}ttacks on \underline{F}ine-\underline{T}uned models (BAFT) are limited by strong assumptions, where their settings assume that the target model is fine-tuned from a source model, associated with some additional knowledge.
In \cite{wang_attackTL_usenix_2018}, it is assumed that the parameters of the low-level layers in the target model are copied from the source model and remain unchanged during the fine-tuning process. But in practice, all the parameters in the target model might be adapted during fine-tuning. The assumption about the training process is relaxed in \cite{zhang_attackTL_kdd_2020} by allowing fine-tuning the whole network, but it is assumed that the network architecture of the source model is known, and they introduce an auxiliary domain that shares the same input and output spaces as those of the target domain to craft adversarial examples. The knowledge about the network architecture might be unavailable to the attacker, and the introduction of the auxiliary domain requires additional data labeling costs, which limit the feasibility of the proposed attack method. 

\begin{figure*} 
\centering
    \begin{minipage}{0.48\linewidth}
    \centering
    \subfloat[Cross-domain BAFT (CD-BAFT)] {
    \centering
    \includegraphics[scale=0.25]{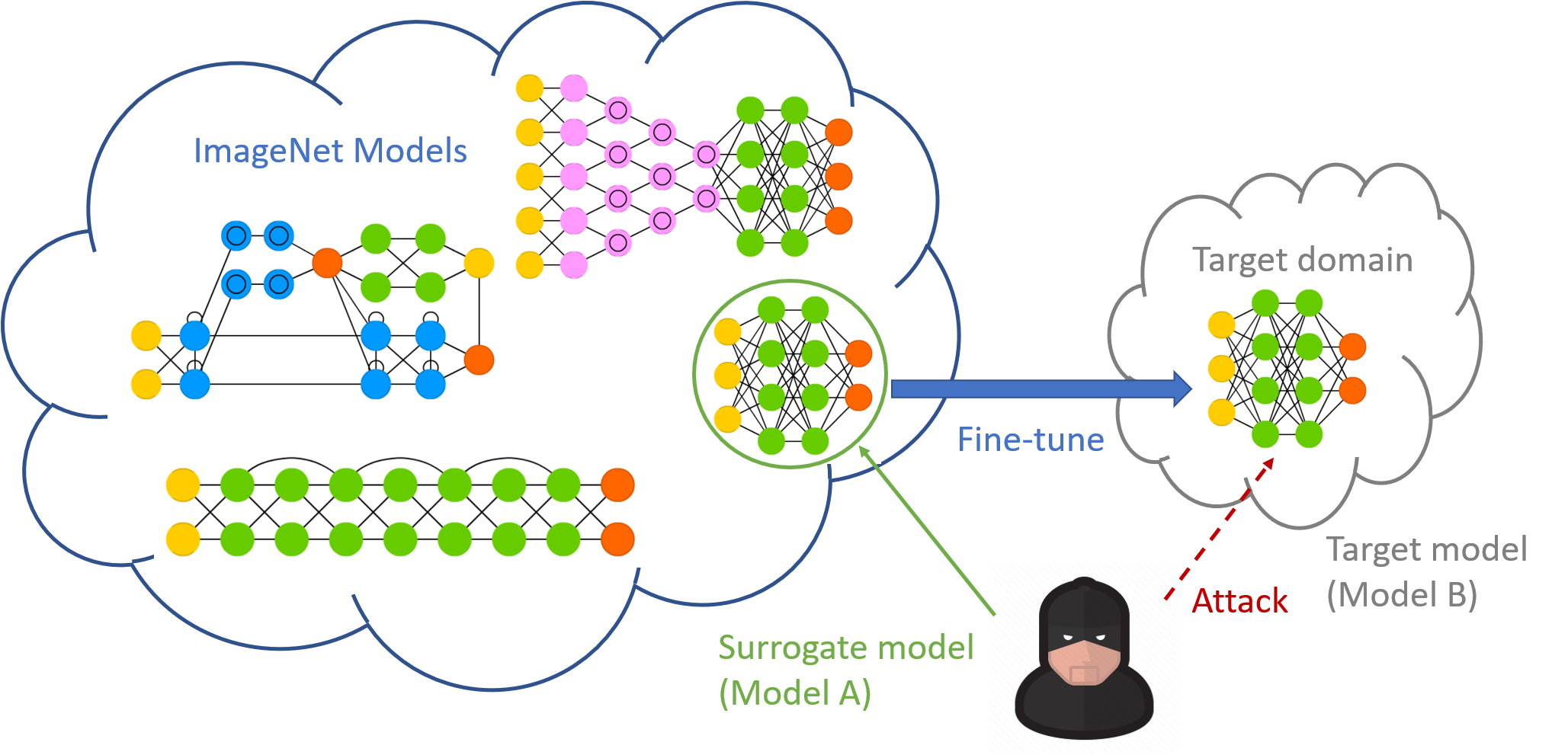}
    \label{fig:cross-domain-baft}
    }
    \end{minipage}
    \hfill
    \begin{minipage}{0.48\linewidth}
    \centering
    \subfloat[Cross-domain cross-architecture BAFT (CDCA-BAFT)] {
    \centering
    \includegraphics[scale=0.25]{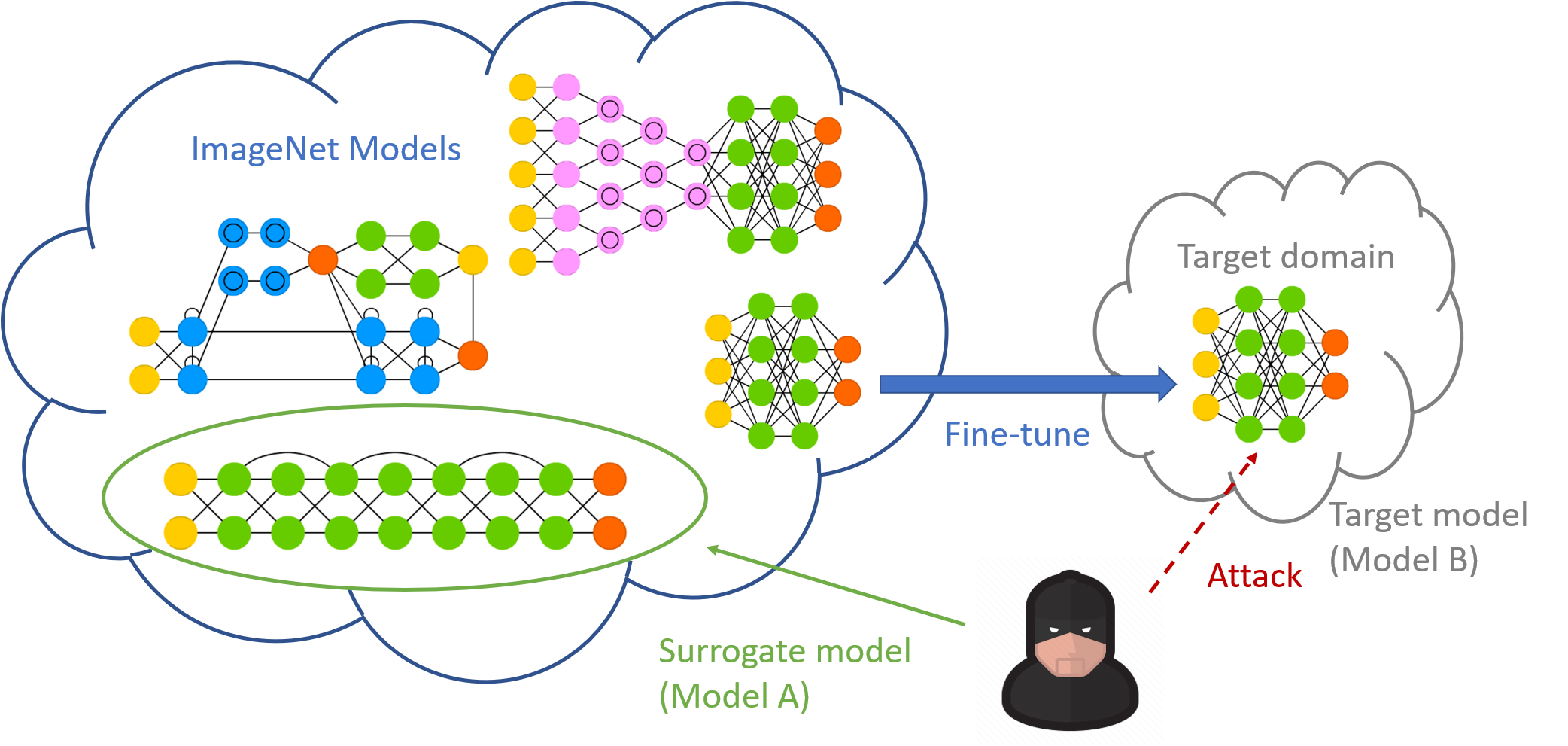}
    \label{fig:cross-arch-baft}
    }
    \end{minipage}
    \caption{
    A schematic view of the two BAFT settings. \texttt{Model} \texttt{A} and \texttt{Model} \texttt{B} can have different architectures, which corresponds to two scenarios, cross-domain BAFT (CD-BAFT) and cross-domain cross-architecture BAFT (CDCA-BAFT). 
    If the attacker coincidentally chooses the exact source model that the target model is fine-tuned from, it corresponds to the CD-BAFT setting, as shown in (a). If the attacker uses a source model whose architecture is different from that of the target model, it corresponds to the CDCA-BAFT setting, as shown in (b). 
    Though there are discrepancies in terms of both domain and network architecture, effective adversarial attacks towards \texttt{Model} \texttt{B} can be achieved with the help of \texttt{Model} \texttt{A}. 
    }
    \label{fig:two-baft-settings}
\end{figure*}

Thus, we study two novel but more difficult BAFT settings: \emph{cross-domain} (CD) and \emph{cross-domain cross-architecture} (CDCA) BAFT, which assume that (1) the target model is a fine-tuned model, and (2) the source domain data is known and accessible. 
The CD-BAFT setting assumes that the network architecture of the source model and that of the target model are the same while the CDCA-BAFT setting further relaxes the assumption by allowing different architectures. For example, the target model for attacking is a VGG-16-based product image classifier, and it is fine-tuned from a VGG-16 model that is pre-trained on ImageNet. If the attacker coincidentally chooses the exact source model that the target model is fine-tuned from, it corresponds to the CD-BAFT setting. In the CDCA-BAFT setting, the attacker might use a ResNet-50-based ImageNet classifier to attack the target model. The differences between the proposed BAFT settings and the vanilla black-box adversarial attack setting are illustrated in Fig. \ref{fig:schema-baft}. A schematic view of the two BAFT settings is provided in Fig.~\ref{fig:two-baft-settings}. 

The goal of the CD- and CDCA-BAFT settings is to generate adversarial examples that successfully fool the target model using limited queries with the knowledge about the source domain. 
Existing black-box attack methods, which can be categorized into two approaches, namely \emph{surrogate-based} and \emph{query-based} methods, cannot be trivially applied to our new BAFT settings. Surrogate-based methods use a substitute model as a white-box model, generate adversarial examples using gradient-based white-box attack methods, and finally attack the target model with the generated examples \cite{papernot_substitute_acm_ccs_2017}. In our cross-domain settings, it is difficult to obtain the surrogate gradient from the substitute model if the source and target domains have different label spaces.
Thus, we cannot compute the gradient via back-propagation using the substitute model. 
Query-based methods estimate the gradient with zeroth-order optimization methods \cite{chen_zoo_acm_ais_2017,ilyas_ql_icml_2018}. Though they estimate gradients without accessing the internal parameters of the target model, large amounts of queries are needed to obtain an accurate gradient estimate, particularly for high-dimensional inputs like images, which makes the attack inefficient. Hence, existing methods cannot meet the requirements of effectiveness and efficiency and the BAFT task remains an open challenge.

To address the above challenges, we propose Transferred Evolutionary Strategies (TES). Our method is composed of two stages. In the first stage, we train an adversarial generator in the source domain. The adversarial generator adopts an encoder-decoder architecture. It first maps the clean input to a low-dimensional latent space and then decodes the latent representation into an adversarial example. The low-dimensional latent space encodes semantic adversarial patterns that are rather transferable~\cite{baluja_atn_aaai_2018,naseer_cross_domain_perturbations_nips_2019,huang_tremba_iclr_2020}. Furthermore, it limits the search in the second stage in a low-dimensional space and improves the query efficiency. In the second stage, the latent representation is used as a starting point. We obtain the surrogate gradient from the source model using a soft labeling mechanism, which represents a target domain label in the source domain label space as the probability score predicted by the source model. Then the search in the latent space is guided with the surrogate gradient using evolutionary strategies \cite{maheswaranathan_guidedES_icml_2019}. The soft labeling mechanism bridges the heterogeneous label spaces and allows obtaining surrogate gradients, which circumvents the limitation of existing surrogate-based methods. Though the surrogate gradient does not perfectly align with the true gradient, they might be correlated and hence using the surrogate gradient improves the efficiency of the search. Experimental results on multiple domains and various backbone network architectures demonstrate the effectiveness and efficiency of the proposed attack method. 

\section{Related Works}
We review related literature on adversarial attacks and defenses and training deep neural networks with soft labels in the following. 

\noindent\textbf{Adversarial Attacks and Defenses}. Adversarial attacks and defenses have drawn attention from the machine learning research community in recent years \cite{yuan_tnnls_attac_survey_2019,li_attack_review_american_stat_2021}. 
Based on the knowledge that the attacker can access, adversarial attacks are categorized into two types, namely white-box and black-box attacks. White-box attacks assume that the attacker has full knowledge of the target model, and the adversarial examples can be found by gradient computation \cite{szegedy_iclr_2013,goodfellow_fgsm_iclr_2014,moosavi_deepfool_cvpr_2016,papernot_jsma_eurosp_2016,kurakin_bim_iclrw_2017,carlini_cw_attack_sp_2017,madry_pgd_iclr_2018}. On the other hand, black-box attacks can only query the output of the target model. The model output can be either probability scores~\cite{chen_zoo_acm_ais_2017,ilyas_ql_icml_2018,li_nattack_icml_2019,cheng_prgf_nips_2019,huang_tremba_iclr_2020} or a single hard label (top-1 prediction)~\cite{brendel_boundary_attack_iclr_2018,cheng_hardlabel_iclr_2019,cheng_signopt_iclr_2020,chen_rays_kdd_2020}. The two black-box attack settings are referred to as \emph{score-based} and \emph{decision-based} attacks. In this paper, we focus on the score-based attacks and leave decision-based attacks for future exploration. 

Score-based black-box attack methods can be categorized into two approaches, namely \emph{surrogate-based} \cite{papernot_substitute_acm_ccs_2017,liu_arch_transfer_iclr_2016} and \emph{query-based} \cite{chen_zoo_acm_ais_2017,ilyas_ql_icml_2018,bhagoji_grad_estimation_eccv_2018,li_nattack_icml_2019,tu_autozoom_aaai_2019} methods. Surrogate-based methods train a substitute model, attack the substitute model with white-box attack methods, and finally attacks the target model using the generated adversarial examples. Query-based methods use gradient-free optimization methods, for example, zeroth-order-optimization \cite{chen_zoo_acm_ais_2017} and evolutionary strategies \cite{ilyas_ql_icml_2018,li_nattack_icml_2019}. Surrogate-based methods cannot be readily applied to the proposed BAFT settings because it is hard to obtain surrogate gradients from the substitute model due to the different label spaces of the two domains. Query-based methods need large amounts of queries, which makes the attack inefficient. 

Recently, there are a few attempts to combine surrogate-based and query-based methods \cite{cheng_prgf_nips_2019,guo_subspace_neurips_2019,huang_tremba_iclr_2020}, and they achieve state-of-the-art attack results on vanilla black-box attacks. The Prior-RGF method proposed in \cite{cheng_prgf_nips_2019} searches adversarial examples in the original input space under the guidance of surrogate gradients. The subspace attack~\cite{guo_subspace_neurips_2019} is conducted in low-dimensional subspaces spanned by surrogate gradients. But these methods cannot be directly applied to the BAFT settings due to the mismatched label spaces. The TREMBA method proposed in \cite{huang_tremba_iclr_2020} optimizes in a transferable low-dimensional latent space, but they do not utilize surrogate gradients, which makes their attack less efficient. Hence, existing black-box attack methods cannot address the unique challenges in the proposed BAFT settings. 

Most adversarial attacks assume that the target model of attack is trained in an individual domain, while attacks on fine-tuned models are less studied. The black-box attack method towards fine-tuned models proposed in \cite{wang_attackTL_usenix_2018} assumes that the target model copies and freezes the first few layers from a pre-trained model, which might not hold in practice. Zhang et al. systematically evaluate the robustness of transfer learning models under both white-box and black-box FGSM attacks in \cite{zhang_attackTL_kdd_2020}. They introduce an auxiliary domain to bridge the label space discrepancy. Their method suffers from two limitations. Firstly, the introduction of the auxiliary domain requires additional data labeling costs. Secondly, they assume that the network architecture of the pre-trained model is known. Compared to these two works, our proposed BAFT settings are more practical by allowing cross-architecture attacks. In addition, our proposed attack method does not require auxiliary domain annotation. 

To defend against adversarial attacks, one natural and effective approach is to augment the training data with adversarial examples, which is known as adversarial training~\cite{madry_pgd_iclr_2018,tramer_ensemble_adv_train_iclr_2018,zhang_trades_icml_2019}. The attack and defense methods compete against each other and formulate an arms race. The initial move to conduct successful attacks in the BAFT settings is crucial for advancing the research in developing robust fine-tuned models. 

\noindent\textbf{Soft Labels}. 
For multi-class classification tasks, data instances usually adopt a ``hard'' labeling mechanism. Each instance is assigned to a class, and the membership is binary, i.e., the instance is either a member of a class or not. Soft labels are defined as a weighted mixture of hard labels and the uniform distribution or probability scores, and they have been used to improve network performance \cite{szegedy_labelsmooth_cvpr_2016} and knowledge distillation \cite{hinton_kd_arxiv_2015}. In this paper, we use soft labels to bridge the gap between the mismatched label spaces of the two domains. We represent target domain labels as the probability scores predicted by the source model and then obtain the surrogate gradient. To the best of our knowledge, this is the first attempt that applies soft labels to cross-domain black-box adversarial attacks. 

\begin{figure*}
    \centering
    \includegraphics[scale=0.9]{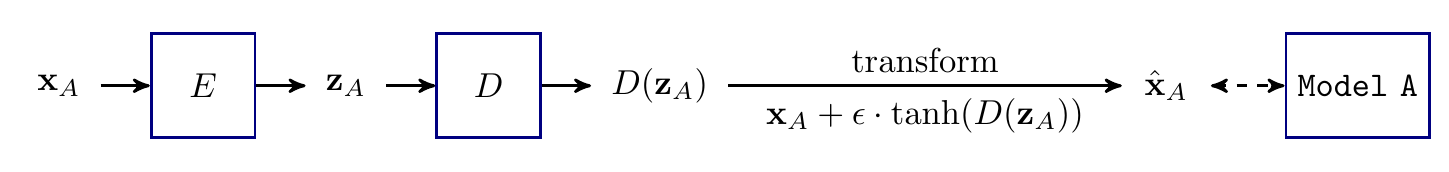}
    \caption{Stage I: training the adversarial generator in the source domain $\mathcal{D}_A$. The adversarial generator adopts an encoder-decoder architecture. It maps a clean example to an adversarial one. The unnormalized output of the decoder is transformed, and there is $\hat{\mathbf{x}}_A = \mathbf{x}_A + \epsilon \cdot \tanh(D(\mathbf{z}_A)) $. It is easy to verify that there is $\lVert \hat{\mathbf{x}}_A - \mathbf{x}_A \rVert_{\infty} \leq \epsilon$ and the adversarial example is ensured to fall within the $\epsilon$-ball of the clean example. }
    \label{fig:tes_stage_I}
\end{figure*}

\section{Transferred Evolutionary Strategies}
We first introduce the problem setup and then present the proposed black-box attack method. Our source code is available at \url{https://github.com/HKUST-KnowComp/TES}.

\begin{figure}
    \centering
    \begin{minipage}{0.45\linewidth}
    \centering
    \subfloat[Transferred Evolutionary Strategies] {
    \centering
    \includegraphics[scale=0.9]{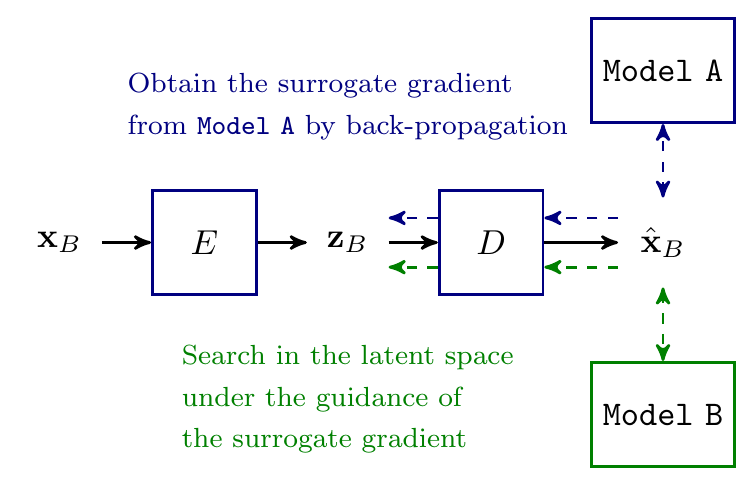}
    \label{fig:schematic-stage-II}
    }
    \end{minipage}
    \hspace{0.1in}
    \begin{minipage}{0.45\linewidth}
    \centering
    \subfloat[A schematic view of guided evolutionary strategies] {
    \centering
    \includegraphics[scale=0.25]{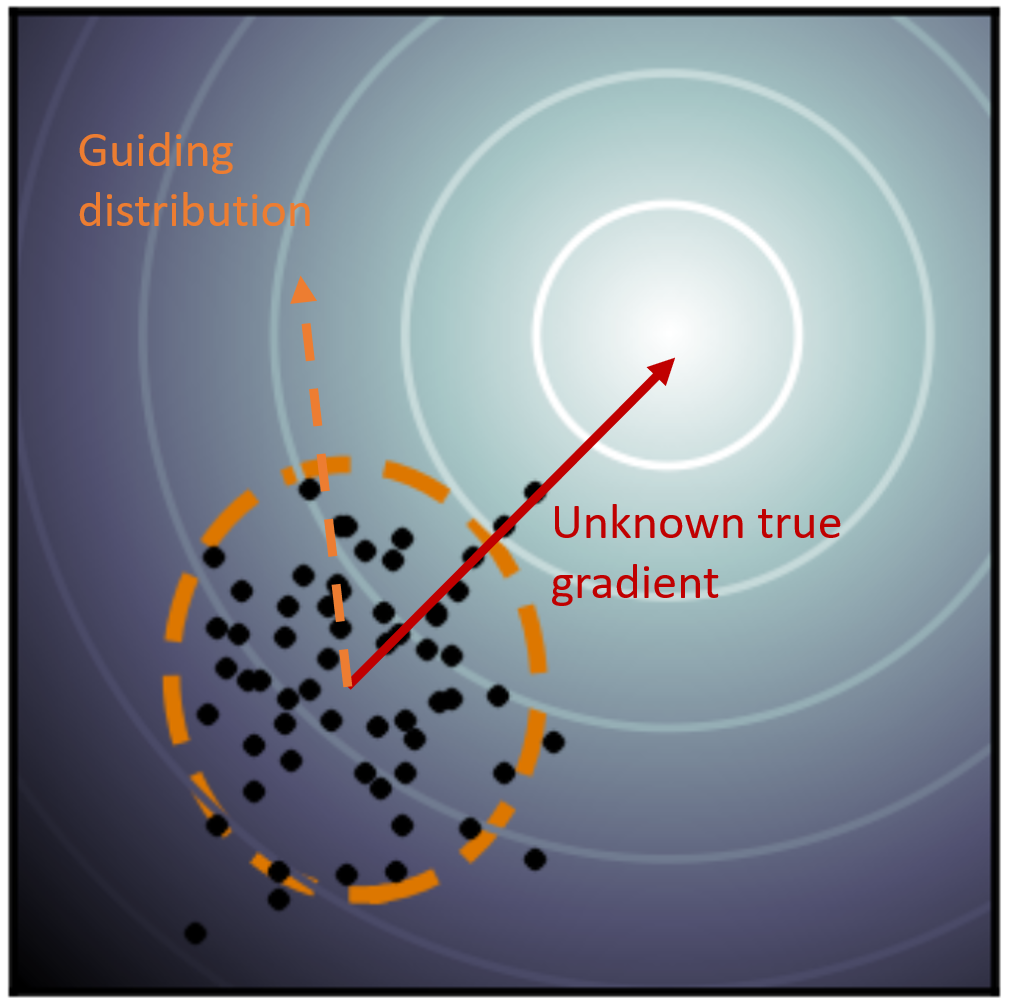}
    \label{fig:schematic-guidedes}
    }
    \end{minipage}
    \caption{Stage II: searching in the latent space under the guidance of the surrogate gradient. Though the surrogate gradient (the orange dashed arrow in (b)) does not perfectly align with the true gradient (the red arrow), it provides useful supervision and accelerates the search. }
    \label{fig:tes_stage_II}
\end{figure}
\subsection{Problem Setup}
We focus on the black-box attacks on the image classification models in this paper. The target model for attacking is a fine-tuned model. To avoid confusion with the target model for attacking, the source domain and the target domain are denoted by $\mathcal{D}_A$ and $\mathcal{D}_B$, respectively. Without the loss of generality, we omit the subscript that indicates the domain. Let $n$ denote the number of labeled samples in a domain, and there is $\mathcal{D} = \{(\mathbf{x}_1, y_1), \ldots, (\mathbf{x}_n, y_n)\}$. The input and output spaces are denoted by $\mathcal{X}$ and $\mathcal{Y}$, respectively. A neural network, denoted by $f$, learns the mapping from the input space to the output space, and there is $f: \mathcal{X} \rightarrow \mathcal{Y}$. The image classifier is usually a Convolutional Neural Network (CNN). The output of the network, denoted by $f(\mathbf{x})$, predicts the probability over the whole label space. 

To attack a neural network $f$, an adversarial example, denoted by $\hat{\mathbf{x}}$, can be found by solving the following optimization problem: 
\begin{equation*}
    \argmin_{\lVert \hat{\mathbf{x}} - \mathbf{x} \rVert_{p} \leq \epsilon} \ell (f(\hat{\mathbf{x}}), \hat{y}), 
\end{equation*}
where $\lVert \cdot \rVert_{p}$ denotes the $p$-norm distance, $\epsilon$ denotes the perturbation budget, $\ell(\cdot, \cdot)$ denotes a classification loss, and $\hat{y}$ denotes a label that is different from the ground truth label $y$ ($\hat{y} \neq y$). The objective of the above optimization problem is to find an adversarial example within the $p$-norm $\epsilon$-ball of the clean input $\mathbf{x}$ that makes the network output an incorrect prediction $\hat{y}$. We assume that the classification loss is the cross-entropy loss, and the distance measure is the $\ell_{\infty}$ distance in the following. There are two types of adversarial attacks, namely \emph{untargeted} and \emph{targeted} attacks. For untargeted attacks, an arbitrary label that is different from the ground truth label is considered a successful attack. For targeted attacks, $\hat{y}$ should equal to a pre-defined label, denoted by $y_t$, where there is $y_t \neq y$. 

The BAFT settings assume that the target model is fine-tuned from a source model. Hence $f_B$ is fine-tuned from a model trained in $\mathcal{D}_A$. The CD-BAFT setting assumes that $f_B$ is directly fine-tuned from $f_A$, while $f_A$ and $f_B$ can have different architectures in the CDCA-BAFT setting. The two settings are illustrated in Figs. \ref{fig:cross-domain-baft} and \ref{fig:cross-arch-baft}, respectively. 

\subsection{Transferred Evolutionary Strategies}
We propose a two-stage black-box attack method that leverages the prior knowledge from the source domain. In the first stage, we train an adversarial generator that maps a clean image to an adversarial one. In the second stage, we obtain a surrogate gradient from the source model with the soft label mechanism that handles the heterogeneous label spaces of the two domains. Then the surrogate gradient is exploited to guide the search in the latent feature space parameterized by the adversarial generator. The details of the adversarial generator training, the soft labeling mechanism, and the guided evolutionary search are given in subsequent sections. The illustrations of the two stages are shown in Figs.~\ref{fig:tes_stage_I} and \ref{fig:schematic-stage-II}. 

\subsubsection{Adversarial Generator Training}
Let $G$ denote the adversarial generator. It maps a clean example to an adversarial one: $G(\mathbf{x}) = \hat{\mathbf{x}}$. The adversarial generator is composed of an encoder and a decoder which are denoted by $E$ and $D$, respectively. There is $G = E \circ D$. We denote the latent representation produced by the encoder by $\mathbf{z}$ and there is $\mathbf{z} = E(\mathbf{x})$. The decoder then projects the latent representation into an unnormalized perturbation. To ensure that the adversarial example falls within the $\epsilon$-ball of the clean input, the unnormalized perturbation is firstly transformed with a $\tanh$ function and then scaled by a factor $\epsilon$: $\hat{\mathbf{x}} = \mathbf{x} + \epsilon \cdot \tanh(D(\mathbf{z})) $. Hence the condition $\lVert \hat{\mathbf{x}} - \mathbf{x} \rVert_{\infty} \leq \epsilon$ is satisfied. 

To train the adversarial generator, we adopt the C\&W attack loss \cite{carlini_cw_attack_sp_2017}. The loss function for untargeted and targeted attacks are defined in Eqs. (\ref{eqn:ag_untargeted}) and (\ref{eqn:ag_targeted}), respectively. 
\begin{equation}
\small{
    \mathcal{L}_{untargeted}(\mathbf{x}, y; f) = \max(F(\hat{\mathbf{x}})_{y} - \max_{k \neq y} F(\hat{\mathbf{x}})_{k}, -\delta), 
}
\label{eqn:ag_untargeted}
\end{equation}
where $F(\mathbf{x})_{j}$ denotes the $j$-th dimension of the output of the classifier $f$ before the Softmax transformation, and $\delta$ denotes a threshold. By optimizing the loss function, the logit value at the $k$-th dimension is expected to be larger than the logit value at the ground truth label by the margin $\delta$ so that the network will make an incorrect prediction other than the ground truth label $y$. 

Similarly, the loss function for targeted attacks is: 
\begin{equation}
\small{
    \mathcal{L}_{targeted}(\mathbf{x}, y; f) = \max(\max_{k \neq y_t} F(\hat{\mathbf{x}})_{k} - F(\hat{\mathbf{x}})_{y_t}, -\delta), 
}
\label{eqn:ag_targeted}
\end{equation}
where $y_t$ is the pre-defined target class. By optimizing $\mathcal{L}_{targeted}$, the classifier $f$ tends to output the prediction $y_t$ with the generated adversarial example as the input. 

\subsubsection{Soft Labeling}
\label{sec:soft_labeling}
In the second stage, we need to search in the low-dimensional latent space parameterized by the encoder under the guidance of the surrogate gradient from the source model $f_A$. 

If the label $y$ is available, the surrogate gradient with respect to the latent representation, denoted by $\nabla_{\mathbf{z}} \mathcal{L}(f_{A}(\hat{\mathbf{x}}), y)$, can be obtained by a simple backward propagation. However, this cannot be trivially applied in the BAFT settings because the label spaces of the source domain and the target domain do not agree, and it is unknown how to represent a target domain label $y_B$ in the source domain label space $\mathcal{Y}_A$. For example, suppose the source domain is ImageNet, and there are two classes, ``batteries'' and ``alarm clock'', in the target domain. For the ``batteries'' class, it does not appear in the ImageNet label space. While for the ``alarm clock'' class, there are multiple ``clock'' related classes in the ImageNet label space, including ``analog clock'', ``digital clock'', and ``wall clock''. There is no correspondence between the two label spaces, and it is also difficult to build one manually. 

To circumvent the heterogeneity of the label spaces, we utilize a soft labeling mechanism in this paper. Let $s_k$ denote the representation of the $k$-th class of the target domain ($k \in \mathcal{Y}_B$) in the source domain label space, and $n_k$ denote the number of samples in the target domain whose label is $k$. Since high-level layers of a CNN contain visual semantic information, the soft label $s_{k}$ is defined as the mean of the probability predictions produced by the source domain network: 
\begin{equation}
    s_{k} = \frac{1}{n_{k}} \sum_{j=1}^{n_k} f_A(\mathbf{x}_j), \quad y_j = k. 
\label{eqn:soft_label}
\end{equation}
Hence we build a mapping between the two label spaces. Consequently, the surrogate gradient can be obtained by $\nabla_{\mathbf{z}} KLD(f_{A}(\hat{\mathbf{x}}), s_{y})$ where $KLD$ denotes the Kullback-Leibler divergence. 

\begin{wrapfigure}{L}{0.55\textwidth}
\begin{minipage}{0.55\textwidth}
\begin{algorithm}[H]
\begin{algorithmic}[1]
\caption{Transferred Evolutionary Strategies}
\label{alg}
  \Require{A clean sample $(\mathbf{x}, y)$, the attack target $f_B$, a source domain $\mathcal{D}_A$, a source model $f_A$, learning rate $\eta$, maximum number of queries $C$, hyper-parameters of guided ES $(\sigma, \alpha, \beta, P)$}.
  \Ensure{An adversarial example $\hat{\mathbf{x}}$. }
  \State Train $G$ with \texttt{Domain} \texttt{A} data to attack $f_A$. 
  \State Compute soft labels using Eq. (\ref{eqn:soft_label}). 
  \State Initialize the latent representation $\mathbf{z}^{(0)} = E(\mathbf{x})$. 
  \State Initialize the adversarial example $\hat{\mathbf{x}}^{(0)} = \mathbf{x} + \epsilon \cdot \tanh(D(\mathbf{z}^{(0)})) $
  \For{$c = 1$ to $C$}
  \State Obtain the surrogate gradient $\nabla_{\mathbf{z}^{(c)}} KLD(f_{A}(\hat{\mathbf{x}}^{(c)}), s_y)$. 
  \State Update the guiding subspace $U$ using Eq. (\ref{eqn:subspace_u}). 
  \State Define the covariance $\Sigma$ using Eq. (\ref{eqn:guided_es_cov_matrix}). 
  \For{$i = 1$ to $P$}
  \State Sample a noise vector $\nu_{i} \sim \mathcal{N}(0, \sigma^{2} \Sigma)$. 
  \State Query $f_B$ and obtain an antithetic pair of losses.
  \EndFor
  \State Compute the gradient estimate using Eq. (\ref{eqn:guided_es_gradient}). 
  \State Update $\mathbf{z}^{(c)}$: $\mathbf{z}^{(c)} = \mathbf{z}^{(c-1)} - \eta \cdot g$.
  \State Generate an adversarial example: $\hat{\mathbf{x}}^{(c)} = \mathbf{x} + \epsilon \cdot \tanh(D(\mathbf{z}^{(c)})) $.
  \If{$\hat{\mathbf{x}}^{(c)}$ successfully attacks $f_B$}
  \State $\hat{\mathbf{x}} = \hat{\mathbf{x}}^{(c)}$
  \State Break
  \EndIf
  \EndFor
\end{algorithmic}
\end{algorithm}
\end{minipage}
\end{wrapfigure}

\subsubsection{Guided Evolutionary Strategies}
With the latent representation $\mathbf{z}$ produced by the encoder and the surrogate gradient obtained from the pre-trained model, we then search in the low-dimensional latent space under the guidance of the surrogate gradient. To incorporate the prior knowledge from the surrogate gradient, we adopt the guided evolutionary strategies (guided ES) \cite{maheswaranathan_guidedES_icml_2019}. 

The search is conducted in an iterative manner. We start from the latent representation produced by the encoder, and there is $z^{(0)} = E(\mathbf{x})$.  The dimension of the latent representation is denoted by $d$. Let $C$ denote the maximum number of queries that is allowed by the target model. At the $c$-th step ($c \in {1, \ldots, C}$), an adversarial example $\hat{\mathbf{x}}^{(c)}$ can be obtained by feed-forwarding the latent representation through the decoder. 
A surrogate gradient, denoted by $\nabla_{\mathbf{z}^{(c)}} KLD(f_{A}(\hat{\mathbf{x}}^{(c)}), s_y)$, can be obtained by back-propagation after querying \texttt{Model} \texttt{A}. Then an orthonormal basis for the subspace, denoted by $U$, can be generated with the surrogate gradient by QR decomposition: 
\begin{equation}
\label{eqn:subspace_u}
U = QR(\nabla_{\mathbf{z}^{(c)}} KLD(f_{A}(\hat{\mathbf{x}}^{(c)}), s_y)), 
\end{equation}
where $U$ is an orthogonal matrix and $QR$ denotes the QR decomposition operation. 

A noise vector, denoted by $\nu$, is sampled from $\mathcal{N}(0, \sigma^{2} \Sigma)$ with: 
\begin{equation}
\label{eqn:guided_es_cov_matrix}
    \Sigma = \frac{\alpha}{d} I_{d} + (1-\alpha) \cdot U U^{T}, 
\end{equation}
where $I_{d}$ denotes a $d \times d$ identity matrix and $\alpha$ is a hyperparameter that balances the search in the full parameter space and the guiding subspace. 

The gradient can then be estimated by antithetic sampling: 
\begin{equation}
\label{eqn:guided_es_gradient}
    g = \frac{\beta}{\sigma^2 P} \sum_{i=1}^{P} \nu_i \cdot [\mathcal{L}(\hat{\mathbf{x}}^{(c)}_{+}, y; f_B) - \mathcal{L}(\hat{\mathbf{x}}^{(c)}_{-}, y; f_B)], 
\end{equation}
where $P$ denotes the number of noise samples, $\beta$ denotes the overall scale of the estimate, and there are $\hat{\mathbf{x}}^{(c)}_{+} = \mathbf{x} + \epsilon \cdot \tanh(D(\mathbf{z}^{(c)}+\nu_i))$ and $\hat{\mathbf{x}}^{(c)}_{-} = \mathbf{x} + \epsilon \cdot \tanh(D(\mathbf{z}^{(c)}-\nu_i))$. 


To summarize, the proposed black-box attack method is outlined in Algorithm \ref{alg}. The guided ES process is illustrated in Fig.~\ref{fig:schematic-guidedes}. 


\begin{minipage}{\textwidth}
\begin{minipage}{0.4\textwidth}
\centering
    \includegraphics[scale=0.35]{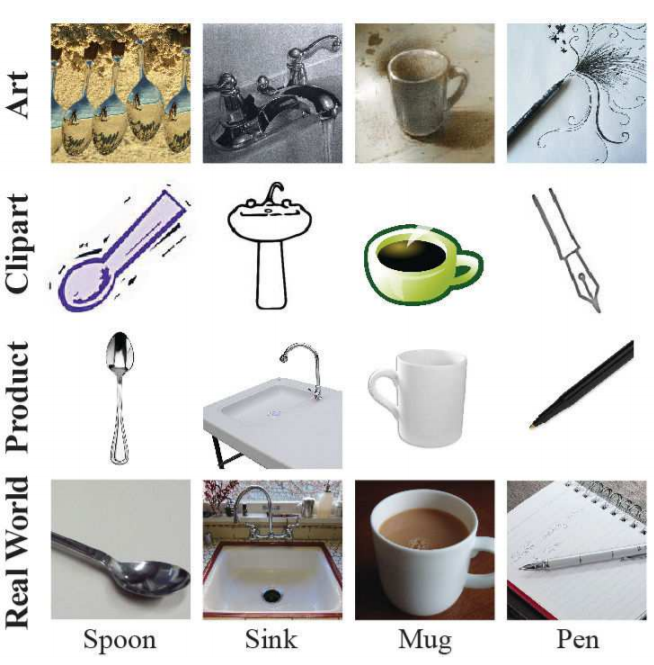}
    \captionof{figure}{Sample images from the \textit{Office-Home} dataset~\cite{venkateswara_office_home_cvpr_2017}. 
    }
    \label{fig:office-home-samples}
\end{minipage}
\hfill
\begin{minipage}{0.55\textwidth}
\centering
\captionof{table}{Dataset Statistics. }
\label{tab:dataset_stat}
\resizebox{\linewidth}{!}{%
\begin{tabular}{ccccc}
\toprule
\multicolumn{2}{c}{Dataset}               & \# classes          & \# train set & \# test set \\
\midrule
\multicolumn{2}{c}{ImageNet}              & 1,000               & 1.28M        & 50K         \\ \hline
\multirow{4}{*}{Office-Home} & \texttt{Art}        & \multirow{4}{*}{65} & 1,941        & 486         \\
                             & \texttt{Clipart}    &                     & 3,492        & 873         \\
                             & \texttt{Product}    &                     & 3,551        & 888         \\
                             & \texttt{Real World} &                     & 3,485        & 872        \\
\bottomrule
\end{tabular}
}
\end{minipage}

\end{minipage}

\section{Experimental Results}
We present our experimental results in the following. 
\begin{table*}[t]
\caption{
Untargeted CD-BAFT attacks. Surrogate-based methods achieve some success without any query since the adversarial examples produced by the source model are transferable. \texttt{AG} provides a good start point for query-based methods. \texttt{TES} improves over \texttt{TREMBA} in terms of the fool rate with reduced queries. 
}
\label{tab:cross-domain-untargeted}
\centering
\begin{tabular}{ccccccc|cc}
\toprule
 \multirow{2}{*}{Domain} &  \multirow{2}{*}{Upper}  &  \multicolumn{5}{c|}{Fool rate}     &   \multicolumn{2}{c}{Mean queries}       \\
 \cline{3-9}
 &  & \texttt{FGSM} & \texttt{PGD}  &   \texttt{AG}   & \texttt{TREMBA} & \texttt{TES} & \texttt{TREMBA} & \texttt{TES} \\
\midrule
\texttt{Art}        & 72.02 & 25.72 & 50.41  & 65.43 &  70.78    & \textbf{71.60}        &  78.08    & \textbf{67.83}        \\
\texttt{Clipart}    & 76.17 & 8.71  & 11.34  & 16.04 & 48.11     & \textbf{48.80}      &  1,021.01    & \textbf{1,018.15}       \\
\texttt{Product}    & 90.65 & 36.26 & 72.75  & 85.59 & 88.63     & \textbf{89.30}        &   65.93   & \textbf{51.33}        \\
\texttt{Real World} & 83.03 & 27.06 & 62.16  & 77.64 & 81.31     & \textbf{81.77}        &   76.02   & \textbf{73.32}       \\
\bottomrule
\end{tabular}
\end{table*}

\begin{table*}[t]
\caption{
Targeted CD-BAFT attacks. Since targeted attacks are more challenging, the fool rates decrease and mean queries increase. \texttt{TES} remains the most effective and efficient among all methods.
}
\label{tab:cross-domain-targeted}
\centering
\begin{tabular}{ccccccc|cc}
\toprule
 \multirow{2}{*}{Domain} &  \multirow{2}{*}{Upper}  &  \multicolumn{5}{c|}{Fool rate}     &   \multicolumn{2}{c}{Mean queries}       \\
 \cline{3-9}
 &  & \texttt{FGSM} & \texttt{PGD}  &   \texttt{AG}   & \texttt{TREMBA} & \texttt{TES} & \texttt{TREMBA} & \texttt{TES} \\
\midrule
\texttt{Art} & 71.55   & 1.46 & 26.57  & 39.54 & 60.88     &   \textbf{64.02}     &   473.90   & \textbf{373.25}       \\
\texttt{Clipart} & 76.79   & 0.23 & 0.00   & 4.69  & 28.37     &  \textbf{29.07}     &   1,509.10   & \textbf{1,498.62}      \\
\texttt{Product} & 90.72   & 0.11 & 5.15   & 0.92  & 48.11     &  \textbf{59.79}     &   1,409.73   & \textbf{1,261.27}       \\
\texttt{Real World} & 83.04   & 0.81 & 19.16  & 41.11 & 66.67     &  \textbf{72.47}      &  591.77    & \textbf{450.01}     \\
\bottomrule
\end{tabular}
\end{table*}


\noindent\textbf{Datasets. } ImageNet \cite{ILSVRC_ijcv_2015}, which is a large-scale image database, is used as the source domain. The source domain network $f_A$ is pre-trained on its train set, and the adversarial generator is trained on its validation set. Four transfer tasks are constructed from the \textit{Office-Home} dataset \cite{venkateswara_office_home_cvpr_2017} which is a popular transfer learning benchmark dataset. The dataset statistics are listed in Table \ref{tab:dataset_stat}. There are only a few thousand labeled samples in the training sets of the target domains, and consequently, it is impossible to train deep learning models in the target domains without transferring the knowledge from the source domain. Some sample images from the \textit{Office-Home} dataset are shown in Fig.~\ref{fig:office-home-samples}. The distributions of the four domains are different, and the distribution of an arbitrary target domain is different from that of the ImageNet dataset as well. 

\noindent\textbf{Baselines. }As a novel black-box attack setting, there are no readily available baselines. We adapt the attack method proposed in \cite{zhang_attackTL_kdd_2020}, consider the state-of-the-art attack method on attacking ImageNet classifiers, and compare the proposed attack method with them. 
\begin{itemize}[leftmargin=*]
    \item Surrogate-based methods: Adversarial examples are produced using \texttt{Model} \texttt{A} and directly attack the target model with the generated adversarial examples. Two gradient-based methods, \texttt{FGSM} and \texttt{PGD}, are used. These two methods adapt the method proposed in \cite{zhang_attackTL_kdd_2020} to the BAFT settings by using the soft labeling mechanism presented in Section~\ref{sec:soft_labeling}. We also include attacking with the adversarial generator, denoted by \texttt{AG}, as a baseline. These three methods do not query the target model. 
    \item Query-based method: Some state-of-the-art black-box attack methods~\cite{cheng_prgf_nips_2019,guo_subspace_neurips_2019} on attacking ImageNet classifiers do not apply to the BAFT settings because surrogate gradients cannot be obtained under the mismatched label spaces. We consider \texttt{TREMBA}~\cite{huang_tremba_iclr_2020} that does not rely on surrogate gradients and achieves competing results on attacking ImageNet. 
\end{itemize}

\noindent\textbf{Evaluation Metrics. }The efficacy of the attack method is measured by \emph{fool rate}, which is the proportion of the test set images that successfully attack the target model. \emph{The higher the fool rate is, the more effective the attack is. }For query-based methods, we also report \emph{mean queries} to evaluate the efficiency of the attack. \emph{Smaller mean queries indicate a more efficient attack. }Only the samples that are correctly classified by the target model are used for attacking, and hence the classification accuracy on the clean test set is the upper bound of the fool rate. The best result of each task is highlighted in boldface. 
\begin{table*}[htbp]
\centering
\caption{
Untargeted CDCA-BAFT attacks. Since there are both domain and network architecture discrepancies, the performance degrades compared to the results of untargeted CD-BAFT attacks. 
}
\label{tab:cross-arch-untargeted}
\begin{tabular}{ccccccc|cc}
\toprule
 \multirow{2}{*}{Domain} &  \multirow{2}{*}{Upper}  &  \multicolumn{5}{c|}{Fool rate}     &   \multicolumn{2}{c}{Mean queries}       \\
 \cline{3-9}
 &  & \texttt{FGSM} & \texttt{PGD}  &   \texttt{AG}   & \texttt{TREMBA} & \texttt{TES} & \texttt{TREMBA} & \texttt{TES} \\
\midrule
\texttt{Art} & 72.02   & 9.67 & 9.05   & 20.58 & 66.67     &   \textbf{69.34}     &   420.39   & \textbf{301.44}        \\
\texttt{Clipart} & 76.17   & 6.87  & 4.70   & 8.25  & 44.10     &  \textbf{45.36}     &   1,182.17   & \textbf{1,157.17}       \\
\texttt{Product} & 90.65   & 26.46 & 17.00  & 43.02 & 84.57     &   \textbf{87.05}     &   371.45   & \textbf{227.24}        \\
\texttt{Real World} & 83.03   & 16.40 & 13.76  & 36.58 & 75.46     &  \textbf{79.01}      &  443.10    & \textbf{285.99}       \\
\bottomrule
\end{tabular}
\end{table*}

\begin{table*}[htbp]
\centering
\caption{
Targeted CDCA-BAFT attacks. All surrogate-based attack methods fail in this setting. \texttt{TES} achieves higher fool rates at the cost of fewer queries. 
}
\label{tab:cross-arch-targeted}
\begin{tabular}{ccccccc|cc}
\toprule
 \multirow{2}{*}{Domain} &  \multirow{2}{*}{Upper}  &  \multicolumn{5}{c|}{Fool rate}     &   \multicolumn{2}{c}{Mean queries}       \\
 \cline{3-9}
 &  & \texttt{FGSM} & \texttt{PGD}  &   \texttt{AG}   & \texttt{TREMBA} & \texttt{TES} & \texttt{TREMBA} & \texttt{TES} \\
\midrule
\texttt{Art} &  71.55   & 0.21 & 0.42   & 0.00 & 25.31     &  \textbf{36.40}     &   1,687.60   & \textbf{1,511.83}        \\
\texttt{Clipart} & 76.79   & 0.00 & 0.00   & 0.23 & 12.43     &  \textbf{12.90}      &  1,910.46   & \textbf{1,894.30}      \\
\texttt{Product} & 90.72   & 0.11 & 0.11   & 0.00 & 23.60     & \textbf{ 32.99}     &  1,841.39    & \textbf{1,724.56}       \\
\texttt{Real World} & 83.04   & 0.46 & 0.23   & 0.35 & 27.76     &  \textbf{39.49}     &  1,736.76    & \textbf{1,509.51}       \\
\bottomrule
\end{tabular}
\end{table*}
\subsection{Implementation Details}
All the experiments are implemented with the PyTorch \cite{paszke_pytorch_nips_2019} deep learning framework. 

The target models use \texttt{VGG16} \cite{simonyan_vgg_arxiv_2014}
as backbone networks. The source models are also \texttt{VGG16} networks in the CD-BAFT setting, and they adopt the \texttt{Resnet-50} \cite{he_resnet_cvpr_2016} architecture in the CDCA-BAFT attacks. The backbone networks are pre-trained on ImageNet, and they are provided by the \texttt{torchvision} package. 

The perturbation budget $\epsilon$ equals $8/255$. For the \texttt{PGD} attack, it runs $40$ iterations with a step size of $2/255$. For the query-based attack methods, the maximum number of queries allowed is $2,100$, the number of noise samples $P$ is $20$ and the hyper-parameters $\alpha$, $\beta$, and $\sigma$ are selected from $\{0.1, 0.5, 0.75\}$, $\{1,2\}$, and $\{0.1,1\}$, respectively. 
For the targeted attacks, we set the target class as the ``TV'' class. The images whose labels are ``TV'' are excluded from the test set. For the other images, the objective of the targeted attack is to make the target model output ``TV'' as predictions. 

\subsection{Results}
\label{sec:main_results}
The results of untargeted and targeted CD-BAFT attacks are shown in Tables \ref{tab:cross-domain-untargeted} and \ref{tab:cross-domain-targeted}, respectively. 

For untargeted attacks, the adversarial examples produced by the surrogate-based methods are rather transferable. For example, \texttt{AG} achieves a fool rate at $85.59\%$ on the \texttt{Product} task, which is close to the upper bound of $90.65\%$, and it is achieved without any access to the target domain data or model. \texttt{FGSM} performs the worst among the three surrogate-based methods since it takes only one step along the gradient direction to generate the adversarial example. \texttt{PGD} improves over \texttt{FGSM} by taking iterative steps. \texttt{AG} achieves the highest fool rate among the three surrogate-based methods, and it provides a good starting point for searching. The query-based methods, both \texttt{TREMBA} and \texttt{TES}, outperform the surrogate-based methods by a large margin. Compared to \texttt{TREMBA}, \texttt{TES} further raises the fool rate since it utilizes the surrogate gradient from the source model while \texttt{TREMBA} ignores such prior knowledge. Moreover, the improvement of the fool rate is achieved with reduced mean queries. 

Targeted CD-BAFT attacks are more challenging since the fool rates of all methods decrease. Surrogate-based methods perform poorly on targeted attacks. For example, the fool rates of \texttt{FGSM} are below $2\%$ in all four domains. Query-based methods are again more advantageous than surrogate-based methods. For example, \texttt{AG} almost fails to attack the target model in the \texttt{Product} domain while the fool rate of the \texttt{TREMBA} raises to $48.11\%$, and \texttt{TES} further achieves an improvement of $11.68\%$ compared to \texttt{TREMBA}. 
\begin{table}[htb]
\centering
    \caption{ 
    Untargeted and targeted CD-BAFT attack results in the original input space and the latent space parameterized by the encoder of the adversarial generator. Searching in the low-dimensional latent space noticeably improves the effectiveness and efficiency of the attacks, and the introduction of the adversarial generator is necessary. 
    }
    \label{tab:ablation-dim}
\begin{tabular}{cccccc}
\toprule
 \multirow{2}{*}{Attack type}  &  \multirow{2}{*}{Upper}  & \multicolumn{2}{c}{Fool rate}             & \multicolumn{2}{c}{Mean queries}      \\
 \cline{3-6}
 &  & w/o AG   & \texttt{TES} & w/o AG & \texttt{TES} \\  \midrule
Untargeted & 72.02 & 51.85 &  \textbf{71.60}      &   989.16   & \textbf{67.83}        \\
Targeted & 71.55 & 8.79 &  \textbf{64.02}  &  1,967.25    & \textbf{373.25}       \\ \bottomrule
\end{tabular}
\end{table}

The results of the untargeted and targeted CDCA-BAFT attack results are listed in Tables \ref{tab:cross-arch-untargeted} and \ref{tab:cross-arch-targeted}, respectively. Similar to the results of the CD-BAFT attacks, \texttt{TES} achieves the highest fool rates among all attack methods on all the tasks, and it requires fewer mean queries than \texttt{TREMBA} does. 

For untargeted attacks, the fool rates of the CDCA-BAFT attacks decrease compared to their counterparts of the CD-BAFT attacks. For example, the fool rate of \texttt{AG} on the \texttt{Art} domain is $20.58\%$, which lags behind the fool rate of the untargeted CD-BAFT attack by $44.85\%$. This indicates that the latent representations produced by the adversarial generator are less transferable if the source model and the target model adopt different network architectures. 

Targeted CDCA-BAFT attacks are the most challenging setting since the fool rates of surrogate-based methods are close to $0\%$, and the mean queries of both query-based methods exceed $1,000$. The improvement of \texttt{TES} over \texttt{TREMBA} in terms of the fool rate on the four tasks is $11.09\%$, $0.47\%$, $9.39\%$, and $11.73\%$, respectively. The advantage of \texttt{TES} on the \texttt{Clipart} dataset is not very significant. We hypothesize that this is because the \texttt{Clipart} dataset is the most dissimilar to the source domain ImageNet, as shown in Fig.~\ref{fig:office-home-samples}, and the prior knowledge from the source domain is not informative. 

In summary, the results under different BAFT settings demonstrate that \texttt{TES} is not only effective but efficient as well. 

\begin{table*}[htb]
\centering
\caption{Untargeted and targeted CD-BAFT attacks in the \texttt{Art} domain where the target model uses the \texttt{ResNet-50} network as the backbone network. The proposed \texttt{TES} method is effective for various backbone architectures. }
\label{tab:ablation-arch-cross-domain}
\begin{tabular}{ccccccc|cc}
\toprule
 \multirow{2}{*}{Attack type} &  \multirow{2}{*}{Upper}  &  \multicolumn{5}{c|}{Fool rate}     &   \multicolumn{2}{c}{Mean queries}       \\
 \cline{3-9}
 &  & \texttt{FGSM} & \texttt{PGD}  &   \texttt{AG}   & \texttt{TREMBA} & \texttt{TES} & \texttt{TREMBA} & \texttt{TES} \\
\midrule
Untargeted & 80.04 & 22.02 & 33.13  & 49.38 & 76.34     &   \textbf{78.40}     &    259.76  & \textbf{164.39}       \\
Targeted & 79.71 & 1.26  & 3.35   & 17.78 & 52.93     &   \textbf{58.58}    &   1,064.77   & \textbf{896.51}      \\
\bottomrule
\end{tabular}
\end{table*}

\begin{table*}[htbp]
\centering
\caption{CDCA-BAFT attacks in the \texttt{Art} domain where the target model uses the \texttt{ResNet-50} network as backbone and the source model uses different backbone networks. A more effective attack is achieved with a more expressive backbone network as the source model. }
\label{tab:ablation-arch-cross-arch}
\begin{tabular}{ccccccc|cc}
\toprule
 \multirow{2}{*}{\texttt{Model} \texttt{A} architecture} &  \multirow{2}{*}{Upper}  &  \multicolumn{5}{c|}{Fool rate}     &   \multicolumn{2}{c}{Mean queries}       \\
 \cline{3-9}
 &  & \texttt{FGSM} & \texttt{PGD}  &   \texttt{AG}   & \texttt{TREMBA} & \texttt{TES} & \texttt{TREMBA} & \texttt{TES} \\
\midrule
\texttt{VGG16} & \multirow{2}{*}{80.04} & 13.37 & 12.96  & 12.14 & 60.08     &  \textbf{61.11}      &   827.31   & \textbf{818.95}       \\
\texttt{Densenet161}               &         & 15.02 & 13.79  & 22.63 & 70.37     &  \textbf{72.22}      &  592.93    & \textbf{527.90}   \\
\bottomrule
\end{tabular}
\end{table*}

\section{Ablation Experiments}
Several factors may influence the effectiveness and efficiency of adversarial attacks. We conduct ablation experiments to investigate four factors: (1) the necessity of the adversarial generator, (2) the effectiveness of the soft labels, (3) network architectures, (4) parameter sensitivity.
The ablation experiments are conducted on the \texttt{Art} domain if not specified. 

\subsection{The Necessity of the Adversarial Generator}
In the proposed \texttt{TES} method, we conduct the guided ES in the latent space parameterized by the encoder of the adversarial generator. The guided ES can also be conducted in the original input space, which can be viewed as extending the method in \cite{cheng_prgf_nips_2019} to the BAFT setting. We compare the results obtained with and without the adversarial generator in Table \ref{tab:ablation-dim}. The results demonstrate that searching in the low-dimensional latent space noticeably improves the effectiveness and efficiency of the attacks, and the introduction of the adversarial generator is necessary. 

\subsection{The Effectiveness of Soft Labels}
\begin{table*}[htb]
    \centering
    \caption{ 
    Top 3 predicted classes and their probability scores obtained with the \texttt{VGG16} network trained on ImageNet. Despite the heterogeneous label spaces, the classes predicted by the source model are visually similar to the target domain labels. 
    }
    \label{tab:domain}
    \begin{tabular}{p{0.15\linewidth}p{0.4\linewidth}p{0.4\linewidth}}
    \toprule
    Class names & \texttt{Art} & \texttt{Clipart}  \\
    \midrule
Alarm clock & Analog clock (0.54), stop watch (0.14), wall clock (0.08)    & Analog clock (0.36), wall clock (0.12), stop watch (0.04)            \\
Bed         & Quilt (0.25), studio couch (0.13), four-poster (0.09)        & Envelope (0.10), cradle (0.06), studio couch (0.05)                  \\
Keyboard    & Computer keyboard (0.33), crossword (0.24), space bar (0.18) & Computer keyboard (0.23), space bar (0.19), notebook computer (0.06) \\
    \bottomrule
    \end{tabular}
\end{table*}
As a case study, in Table \ref{tab:domain}, we report the top 3 predictions and their probability scores of three classes in the \texttt{Art} and \texttt{Clipart} domains. The predictions are produced by the ImageNet pre-trained \texttt{VGG16} network. Despite the mismatched label spaces, for the target domain labels, the top predictions in the source domain are some visually similar classes. The qualitative results suggest that the soft labels retain general visual semantic information and might help obtain surrogate gradients. 

\subsection{Network Architectures}
The experimental results in Section \ref{sec:main_results} are obtained when the target models adopt \texttt{VGG16} as the backbone network. We repeat the experiments on the target models whose backbone network is \texttt{ResNet-50}. 

The results of the CD-BAFT setting are shown in Table \ref{tab:ablation-arch-cross-domain}. Similar to the results obtained when the target network is a \texttt{VGG16} network, \texttt{TES} remains effective and efficient, which demonstrates that our proposed method is insensitive to different network architectures. 

The results of untargeted CDCA-BAFT attacks when \texttt{Model} \texttt{A} adopt various network architectures are shown in Table \ref{tab:ablation-arch-cross-arch}. The target model is fixed to use \texttt{ResNet-50} as the backbone network, and the source model adopts two different backbones, \texttt{VGG16} and \texttt{Densenet161}. The attack is more effective when the source model uses a more expressive backbone network as the fool rates of all the methods increase when the source model is a \texttt{Densenet161} network. 

\begin{figure}[htb]
    \begin{minipage}{0.48\linewidth}
    \centering
    \subfloat[The effect on fool rate] {
    \centering
    \includegraphics[scale=0.375]{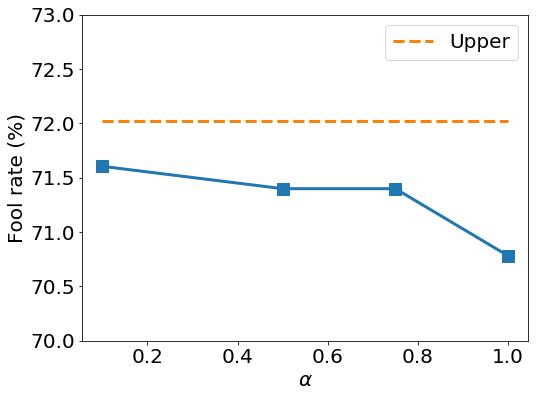}
    }
    \end{minipage}
    \hfill
    \begin{minipage}{0.48\linewidth}
    \centering
    \subfloat[The effect on mean queries] {
    \centering
    \includegraphics[scale=0.375]{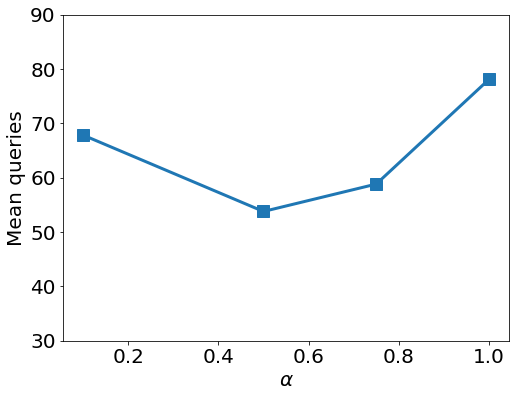}
    }
    \end{minipage}
    \caption{
    Parameter sensitivity of the hyper-parameter $\alpha$, which balances the search from the full parameter space and the guiding subspace. As $\alpha$ increases, the search is biased towards the full parameter space, and the attack becomes less effective and efficient. 
    } 
    \label{fig:ablation-alpha}
    
\end{figure}

\begin{figure}[htb]
    \begin{minipage}{0.48\linewidth}
    \centering
    \subfloat[The effect on fool rate] {
    \centering
    \includegraphics[scale=0.375]{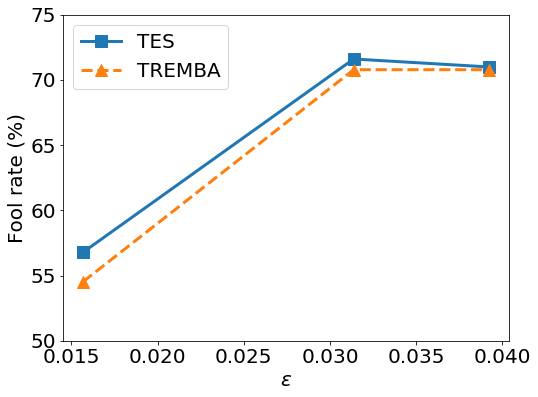}
    }
    \end{minipage}
    \hfill
    \begin{minipage}{0.48\linewidth}
    \centering
    \subfloat[The effect on mean queries] {
    \centering
    \includegraphics[scale=0.375]{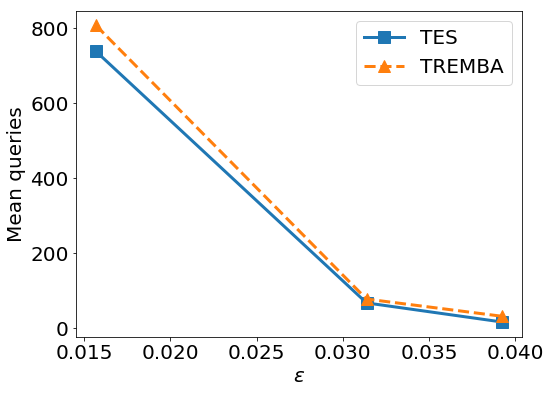}
    }
    \end{minipage}
    \caption{
    Parameter sensitivity of the hyper-parameter $\epsilon$ which controls the perturbation size. The proposed \texttt{TES} method outperforms the \texttt{TREMBA} baseline consistently.
    }
    \label{fig:ablation-eps}
\end{figure}

\subsection{Parameter Sensitivity}
In the guided evolutionary strategies, a hyper-parameter $\alpha$ is introduced to balance the search in the full parameter space and the guiding subspace. We report the fool rates and the mean queries obtained with various $\alpha$ values in Fig. \ref{fig:ablation-alpha}. The lowest fool rate and the maximum mean queries are achieved when there is $\alpha=1.0$, and it corresponds to discarding prior knowledge from the source model and searching in the full parameter space. When there is $\alpha < 1$ and the prior knowledge from the source model is exploited, the effectiveness and efficiency of the attack are improved. 

In addition, we analyze the effect of the perturbation size $\epsilon$ in Fig.~\ref{fig:ablation-eps}. The proposed \texttt{TES} method outperforms the \texttt{TREMBA} baseline consistently under various perturbation sizes. 

\section{Conclusion}
The robustness of fine-tuned neural networks has been less studied despite their prevalence. Thus, we propose two novel cross-domain and cross-domain cross-architecture based BAFT settings. With these two settings, we further propose a two-stage black-box attack method that fails fine-tuned models effectively and efficiently. We expect that the proposed setting and method will facilitate future research on building transfer learning models that are both effective and robust. As future works, we will develop defenses against the attacks in the BAFT settings. 

\begin{ack}
The authors of this paper were supported by the NSFC Fund (U20B2053) from the NSFC of China, the RIF (R6020-19 and R6021-20) and the GRF (16211520) from RGC of Hong Kong, the MHKJFS (MHP/001/19) from ITC of Hong Kong and the National Key R\&D Program of China (2019YFE0198200) with special thanks to HKMAAC and CUSBLT, and  the Jiangsu Province Science and Technology Collaboration Fund (BZ2021065). We also thank the support from the UGC Research Matching Grants (RMGS20EG01-D, RMGS20CR11, RMGS20CR12, RMGS20EG19, RMGS20EG21).
\end{ack}

\bibliographystyle{plain}
\bibliography{references}  

\begin{thebibliography}{10}

\bibitem{baluja_atn_aaai_2018}
Shumeet Baluja and Ian Fischer.
\newblock Learning to attack: Adversarial transformation networks.
\newblock In {\em AAAI}, pages 2687--2695. {AAAI} Press, 2018.

\bibitem{bhagoji_grad_estimation_eccv_2018}
Arjun~Nitin Bhagoji, Warren He, Bo~Li, and Dawn Song.
\newblock Practical black-box attacks on deep neural networks using efficient
  query mechanisms.
\newblock In {\em ECCV}, pages 154--169. Springer, 2018.

\bibitem{brendel_boundary_attack_iclr_2018}
Wieland Brendel, Jonas Rauber, and Matthias Bethge.
\newblock Decision-based adversarial attacks: Reliable attacks against
  black-box machine learning models.
\newblock In {\em ICLR}, 2018.

\bibitem{carlini_cw_attack_sp_2017}
Nicholas Carlini and David Wagner.
\newblock Towards evaluating the robustness of neural networks.
\newblock In {\em IEEE Symposium on Security and Privacy}, pages 39--57. {IEEE}
  Computer Society, 2017.

\bibitem{chen_rays_kdd_2020}
Jinghui Chen and Quanquan Gu.
\newblock Rays: A ray searching method for hard-label adversarial attack.
\newblock In {\em KDD}, pages 1739--1747. {ACM}, 2020.

\bibitem{chen_zoo_acm_ais_2017}
Pin-Yu Chen, Huan Zhang, Yash Sharma, Jinfeng Yi, and Cho-Jui Hsieh.
\newblock Zoo: Zeroth order optimization based black-box attacks to deep neural
  networks without training substitute models.
\newblock In {\em AISec}, pages 15--26. {ACM}, 2017.

\bibitem{cheng_hardlabel_iclr_2019}
Minhao Cheng, Thong Le, Pin{-}Yu Chen, Huan Zhang, Jinfeng Yi, and Cho{-}Jui
  Hsieh.
\newblock Query-efficient hard-label black-box attack: An optimization-based
  approach.
\newblock In {\em ICLR}, 2019.

\bibitem{cheng_signopt_iclr_2020}
Minhao Cheng, Simranjit Singh, Patrick~H. Chen, Pin{-}Yu Chen, Sijia Liu, and
  Cho{-}Jui Hsieh.
\newblock Sign-opt: {A} query-efficient hard-label adversarial attack.
\newblock In {\em ICLR}, 2020.

\bibitem{cheng_prgf_nips_2019}
Shuyu Cheng, Yinpeng Dong, Tianyu Pang, Hang Su, and Jun Zhu.
\newblock Improving black-box adversarial attacks with a transfer-based prior.
\newblock In {\em NeurIPS}, pages 10932--10942, 2019.

\bibitem{goodfellow_fgsm_iclr_2014}
Ian~J Goodfellow, Jonathon Shlens, and Christian Szegedy.
\newblock Explaining and harnessing adversarial examples.
\newblock In {\em ICLR}, 2015.

\bibitem{guo_subspace_neurips_2019}
Yiwen Guo, Ziang Yan, and Changshui Zhang.
\newblock Subspace attack: Exploiting promising subspaces for query-efficient
  black-box attacks.
\newblock In {\em NeurIPS}, pages 3820--3829, 2019.

\bibitem{he_resnet_cvpr_2016}
Kaiming He, Xiangyu Zhang, Shaoqing Ren, and Jian Sun.
\newblock Deep residual learning for image recognition.
\newblock In {\em CVPR}, pages 770--778. {IEEE} Computer Society, 2016.

\bibitem{hinton_kd_arxiv_2015}
Geoffrey Hinton, Oriol Vinyals, and Jeff Dean.
\newblock Distilling the knowledge in a neural network.
\newblock {\em arXiv preprint arXiv:1503.02531}, 2015.

\bibitem{huang_tremba_iclr_2020}
Zhichao Huang and Tong Zhang.
\newblock Black-box adversarial attack with transferable model-based embedding.
\newblock In {\em ICLR}, 2020.

\bibitem{ilyas_ql_icml_2018}
Andrew Ilyas, Logan Engstrom, Anish Athalye, and Jessy Lin.
\newblock Black-box adversarial attacks with limited queries and information.
\newblock In {\em ICML}, volume~80, pages 2142--2151. {PMLR}, 2018.

\bibitem{kurakin_bim_iclrw_2017}
Alexey Kurakin, Ian~J. Goodfellow, and Samy Bengio.
\newblock Adversarial examples in the physical world.
\newblock In {\em ICLR Workshop}, 2017.

\bibitem{li_nattack_icml_2019}
Yandong Li, Lijun Li, Liqiang Wang, Tong Zhang, and Boqing Gong.
\newblock Nattack: Learning the distributions of adversarial examples for an
  improved black-box attack on deep neural networks.
\newblock In {\em ICML}, volume~97, pages 3866--3876. {PMLR}, 2019.

\bibitem{li_attack_review_american_stat_2021}
Yao Li, Minhao Cheng, Cho-Jui Hsieh, and Thomas~CM Lee.
\newblock A review of adversarial attack and defense for classification
  methods.
\newblock {\em The American Statistician}, pages 1--44, 2021.

\bibitem{liu_arch_transfer_iclr_2016}
Yanpei Liu, Xinyun Chen, Chang Liu, and Dawn Song.
\newblock Delving into transferable adversarial examples and black-box attacks.
\newblock In {\em ICLR}, 2017.

\bibitem{madry_pgd_iclr_2018}
Aleksander Madry, Aleksandar Makelov, Ludwig Schmidt, Dimitris Tsipras, and
  Adrian Vladu.
\newblock Towards deep learning models resistant to adversarial attacks.
\newblock In {\em ICLR}, 2018.

\bibitem{maheswaranathan_guidedES_icml_2019}
Niru Maheswaranathan, Luke Metz, George Tucker, Dami Choi, and Jascha
  Sohl-Dickstein.
\newblock Guided evolutionary strategies: augmenting random search with
  surrogate gradients.
\newblock In {\em ICML}, volume~97, pages 4264--4273. {PMLR}, 2019.

\bibitem{moosavi_deepfool_cvpr_2016}
Seyed-Mohsen Moosavi-Dezfooli, Alhussein Fawzi, and Pascal Frossard.
\newblock Deepfool: a simple and accurate method to fool deep neural networks.
\newblock In {\em CVPR}, pages 2574--2582. {IEEE} Computer Society, 2016.

\bibitem{naseer_cross_domain_perturbations_nips_2019}
Muhammad~Muzammal Naseer, Salman~H Khan, Muhammad~Haris Khan, Fahad~Shahbaz
  Khan, and Fatih Porikli.
\newblock Cross-domain transferability of adversarial perturbations.
\newblock In {\em NeurIPS}, pages 12885--12895, 2019.

\bibitem{papernot_substitute_acm_ccs_2017}
Nicolas Papernot, Patrick McDaniel, Ian Goodfellow, Somesh Jha, Z~Berkay Celik,
  and Ananthram Swami.
\newblock Practical black-box attacks against machine learning.
\newblock In {\em ACM ASIACCS}, pages 506--519, 2017.

\bibitem{papernot_jsma_eurosp_2016}
Nicolas Papernot, Patrick McDaniel, Somesh Jha, Matt Fredrikson, Z~Berkay
  Celik, and Ananthram Swami.
\newblock The limitations of deep learning in adversarial settings.
\newblock In {\em EuroS\&P}, pages 372--387. {IEEE}, 2016.

\bibitem{paszke_pytorch_nips_2019}
Adam Paszke, Sam Gross, Francisco Massa, Adam Lerer, James Bradbury, Gregory
  Chanan, Trevor Killeen, Zeming Lin, Natalia Gimelshein, Luca Antiga, et~al.
\newblock Pytorch: An imperative style, high-performance deep learning library.
\newblock In {\em NeurIPS}, pages 8026--8037, 2019.

\bibitem{ILSVRC_ijcv_2015}
Olga Russakovsky, Jia Deng, Hao Su, Jonathan Krause, Sanjeev Satheesh, Sean Ma,
  Zhiheng Huang, Andrej Karpathy, Aditya Khosla, Michael Bernstein,
  Alexander~C. Berg, and Li~Fei-Fei.
\newblock {ImageNet} large scale visual recognition challenge.
\newblock {\em IJCV}, 115(3):211--252, 2015.

\bibitem{simonyan_vgg_arxiv_2014}
Karen Simonyan and Andrew Zisserman.
\newblock Very deep convolutional networks for large-scale image recognition.
\newblock In {\em ICLR}, 2015.

\bibitem{szegedy_labelsmooth_cvpr_2016}
Christian Szegedy, Vincent Vanhoucke, Sergey Ioffe, Jon Shlens, and Zbigniew
  Wojna.
\newblock Rethinking the inception architecture for computer vision.
\newblock In {\em CVPR}, pages 2818--2826. {IEEE} Computer Society, 2016.

\bibitem{szegedy_iclr_2013}
Christian Szegedy, Wojciech Zaremba, Ilya Sutskever, Joan Bruna, Dumitru Erhan,
  Ian Goodfellow, and Rob Fergus.
\newblock Intriguing properties of neural networks.
\newblock In {\em ICLR}, 2014.

\bibitem{tramer_ensemble_adv_train_iclr_2018}
Florian Tram{\`{e}}r, Alexey Kurakin, Nicolas Papernot, Ian~J. Goodfellow, Dan
  Boneh, and Patrick~D. McDaniel.
\newblock Ensemble adversarial training: Attacks and defenses.
\newblock In {\em ICLR}, 2018.

\bibitem{tu_autozoom_aaai_2019}
Chun-Chen Tu, Paishun Ting, Pin-Yu Chen, Sijia Liu, Huan Zhang, Jinfeng Yi,
  Cho-Jui Hsieh, and Shin-Ming Cheng.
\newblock Autozoom: Autoencoder-based zeroth order optimization method for
  attacking black-box neural networks.
\newblock In {\em AAAI}, pages 742--749. {AAAI} Press, 2019.

\bibitem{venkateswara_office_home_cvpr_2017}
Hemanth Venkateswara, Jose Eusebio, Shayok Chakraborty, and Sethuraman
  Panchanathan.
\newblock Deep hashing network for unsupervised domain adaptation.
\newblock In {\em CVPR}, pages 5385--5394. {IEEE} Computer Society, 2017.

\bibitem{wang_attackTL_usenix_2018}
Bolun Wang, Yuanshun Yao, Bimal Viswanath, Haitao Zheng, and Ben~Y Zhao.
\newblock With great training comes great vulnerability: Practical attacks
  against transfer learning.
\newblock In {\em {USENIX} Security}, pages 1281--1297. {USENIX} Association,
  2018.

\bibitem{yuan_tnnls_attac_survey_2019}
Xiaoyong Yuan, Pan He, Qile Zhu, and Xiaolin Li.
\newblock Adversarial examples: Attacks and defenses for deep learning.
\newblock {\em IEEE TNNLS}, 2019.

\bibitem{zhang_trades_icml_2019}
Hongyang Zhang, Yaodong Yu, Jiantao Jiao, Eric Xing, Laurent El~Ghaoui, and
  Michael Jordan.
\newblock Theoretically principled trade-off between robustness and accuracy.
\newblock In {\em ICML}, volume~97, pages 7472--7482. {PMLR}, 2019.

\bibitem{zhang_attackTL_kdd_2020}
Yinghua Zhang, Yangqiu Song, Jian Liang, Kun Bai, and Qiang Yang.
\newblock Two sides of the same coin: White-box and black-box attacks for
  transfer learning.
\newblock In {\em KDD}, pages 2989--2997. {ACM}, 2020.

\end{thebibliography}

\end{document}